# APPLICATION OF FUZZY RULE BASED SYSTEM FOR HIGHWAY RESEARCH BOARD CLASSIFICATION OF SOILS

Sujatha A[1], L Govindaraju[2] and N Shivakumar[3]

[1]Assistant Professor, Department of Mathematics, RV College of Engineering, Bangalore 59, India.
[2]Professor, Department of Civil Engineering, University Visveswaraya College of Engineering, Bangalore-59, India
[3]Professor, Department of Mathematics, RV College of Engineering, Bangalore 59, India

*ABSTRACT*

*Fuzzy rule-based model is a powerful tool for imitating the human way of thinking and solving uncertainty-related problems as it allows for understandable and interpretable rule bases. The objective of this paper is to study the applicability of fuzzy rule-based modelling to quantify soil classification for engineering purposes by qualitatively considering soil index properties. The classification system of the Highway Research Board is considered to illustrate a fuzzy rule-based model. The soil's index properties are fuzzified using triangular functions, and the fuzzy membership values are calculated. Fuzzy arithmetical operators are then applied to the membership values obtained for classification. Fuzzy decision tree classification algorithm is used to derive fuzzy if-then rules to quantify qualitative soil classification. The proposed system is implemented in MATLAB. The results obtained are checked and the implementation of the proposed model is measured against the outcomes of the laboratory tests.*

## 1. INTRODUCTION

Machine learning is a systematic computer program that provides problem solving at the level of a human expert, based on task-specific expertise and inference techniques [1]. These techniques are derived from the analysis of artificial intelligence, a user-friendly interactive computer program that incorporates the skill of an expert or a group of experts in a well-defined domain. The data used in different models of judgment is influenced by unpredictability which leads to uncertainty. Due to the non-descriptive context of social and natural attributes, a primary source of unpredictability is the variance of the results. The alternative type of unpredictability is indefinite, which may be due to the values derived from a measuring instrument or from the observer who is performing the task [2]. Fuzzy expert system (FES) has been shown to tackle such unpredictability to solve the dynamic problems of the real world [3].

Fuzzy expert system is one such system introduced by Kandel (1992) [4]. This is an intelligent tool capable of making decisions and also deals with ambiguous data. FES has improved the excellence, effectiveness and quality in recent times. FES has been developed for numerous real world problems such as numerical classification of soil and mapping, land evaluation, slope stability, rock engineering, tunnelling, project management, wastewater treatment, online scheduling, performance indexing, computer security, gesture recognition, medical diagnosis, agricultural problem to deal the vagueness by mimicking the human way of thinking [5]. Expert System has been applied in soil classification for agricultural purposes where in computer-aided soil classification was developed involving substantial number of rules, inter-parametric associations and subjective assessments [6] [7] [8] . A computer algorithm for the rule based inference process was developed [9], and the application of fuzzy logic, a representation of subjective evaluations was obtained by the application of fuzzy logic [10] [11] provided a representation of subjective evaluations [12].

In 1999, Fetz et al. discussed the application of fuzzy models in geotechnical engineering based on the α-cut set interval analysis [13] . Hayo M.G. et.al, in 1998 proposed a fuzzy expert system or calculating an "Ipest" index that contemplates expert insight into potential environmental impacts through the application of pesticides in the field [14].

Many other applications on FES includes: tunnel boring machine performance modelling [15], prediction of liquefaction [16], model footing response analysis [17], compact soil swelling potential [18], modelling of soil shearing resistance angle using soft computing systems[19].
In 2011 Adoko analyzed applications in geotechnical engineering focused on fuzzy inference technique [20]. Mayadevi, N et al. published an overview of various expert system implementations in power plants in 2014 [21]. In 2014, T.S.Umesha., et al, developed a fuzzy model for parameters of contaminated soil [22].

The goal of this study is to develop an interactive, user-friendly fuzzy rule-based system using fuzzy IF-THEN rules to quantify the soil classification for engineering purposes in qualitative terms, taking into account the soil index properties. Highway research board classification system is considered for the demonstration of fuzzy rule-based model.

The soil's index properties are fuzzified using triangular functions and the fuzzy membership values are calculated. Then fuzzy algebraic operators are applied to the fuzzy membership values for classification. Fuzzy decision tree classification algorithm is used to derive fuzzy if-then rules to quantify qualitative soil classification. The results are validated and compared with the laboratory test results, a common classification method in order to evaluate the efficacy of the proposed model.

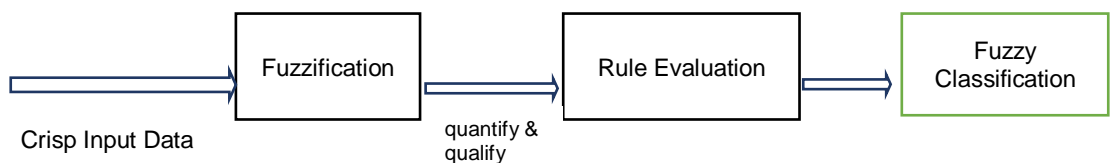

Figure 1: Flow diagram of Fuzzy Rule based

## 1.1. Review of Fuzzy Set Theory

Fuzzy sets, initially introduced by L.A.Zadeh [23], has been applied in numerous fields, such as decision-making and control. Basic definitions of fuzzy sets and fuzzy arithmetic can be found in [24]. A brief review of the fuzzy set's definitions, fuzzy numbers, and fuzzy operations is given below. A fuzzy set is defined as follows.

Definition: Let X be a nonempty set. A fuzzy set Q in X is characterized by its membership function $\mu_Q: X \rightarrow [0,1]$, and $\mu_Q(x)$ is interpreted as the degree of membership of element x in fuzzy set Q for each $x \in X$.

In fuzzy set theory, fuzzy sets are characterized by membership functions [25]. In practice, membership functions are selected arbitrarily. The most widely used membership functions are usually represented in triangular, trapezoidal, Gaussian forms.

The triangular fuzzy number is defined by $\mu_A(x) = \max\left[\min\left(\dfrac{x-a}{b-a}, \dfrac{c-x}{c-b}\right), 0\right]$

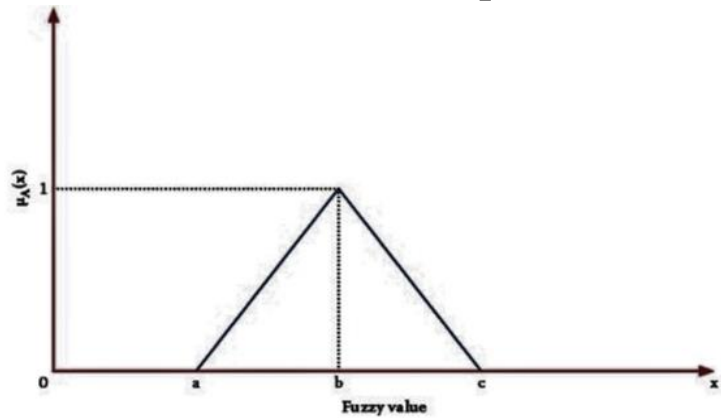

Figure 2: Triangular Fuzzy Number

Rule-based fuzzy classification systems require Fuzzy sets and partitions to granulate the features domain. In the preceding part of the rules, the linguistic variables represent features, and the consequent part is a class. A typical fuzzy classification rule can be expressed as

$R_K$: IF $E_1$ is $P_{1l_1}$ AND $E_2$ is $P_{2l_2}$ AND …. AND $E_m$ is $P_{ml_m}$ THEN Class = $C_i$

where $R_K$ is the rule identifier, $E_1, \ldots\ldots, E_m$ are the features of the example considered in the problem (represented by linguistic variables), $P_{1l_1}, \ldots\ldots\ldots\ldots, P_{ml_m}$ are the linguistic values used to represent the feature values, and $C_i \in C$ is the class.

For a given case, the applicability of a fuzzy rule relies on "grade of truth" or "value of truth", that depends on the reasons to which the rule should be applied. The degree of fulfilment (DOF) of that rule is called the truth value correspondent to the fulfillment of the conditions of law. There are several different methods available for calculating the DOF[1] .

In the current case, a classification should be given by fuzzy rules. An example is categorized by comparing each rule in the Fuzzy rule base to a particular class to which it belongs [26]. Using the fuzzy reasoning process, the number of degrees of compatibility for a set of fuzzy rules for each class is determined and the class with the highest sum is used to classify the particular example considered.

### 1.2. Algorithm to Generate Fuzzy Rules

The hardship lies in finding an optimally working rule structure. The algorithm used to generate fuzzy rules to address the classification problem [27] is as follows.

1) The number of rules for each classification $C_1, C_2, \ldots, C_n$ is selected.
2) Possible fuzzy sets are specified for the input parameters.
3) An initial rule system $R_i$ is developed at random.
4) Calculating the membership values of the input parameters determines the degree of consistency of the rule system.
5) the original rule structure $R_i$ is replaced by a new rule system $R_i^*$.
6) The degree of compatibility is assessed for the new rule system.
7) Steps 4–6 are repeated N number of times and a set of definitive rules is developed.

## 2. HIGHWAY RESEARCH BOARD CLASSIFICATION

Vagueness occurs in the parameters of the data, as their classification relationship is not crisp. In the Fuzzy rule-based model, the input parameters are represented in the form of fuzzy sets. A Fuzzy set is the set of values a parameter can take. The membership function defines all the information contained in a fuzzy package. The parameters are defined in the form of membership functions that cover the likely range of values a parameter will assume in most situations[28]. The inputs are taken through membership functions to discover the degree to which they belong to each one of the respective fuzzy sets. In this analysis, the inputs are fuzzified by a triangular function.

This study aims to build Fuzzy rule-based classification for the HRB classification system in qualitative terms, taking into account the six soil index properties, namely

1. Particle size smaller than 2mm, Particle size smaller than 0.425mm,
2. Particle size smaller than 0.075mm,
3. liquid limit,
4. plastic limit
5. plasticity Index

It is possible to approach the combination of degree of match and the Fuzzy rule-based system for the qualitative HRB soil classification consisting of twelve subgroups as shown in Table 1[29].

Table 1: Subgroups of HRB classification system

| Type | Subgroup | Type | Subgroup |
|------|----------|------|----------|
| 1 | A-1-a | 7 | A-2-7 |
| 2 | A-1-b | 8 | A-4 |
| 3 | A-3 | 9 | A-5 |
| 4 | A-2-4 | 10 | A-6 |
| 5 | A-2-5 | 11 | A-7-5 |
| 6 | A-2-6 | 12 | A-7-6 |

## 2.1. Fuzzy Sets and Membership Functions for Coarse Fraction Passing 2mm

The membership function for particle size smaller than 2mm size is divided into five ranges denoted by fuzzy descriptors, namely Very Low (VL), Low (L), Medium (M), High (H) and Very High (VH) and is defined as follows

$$A_{VL}(p) = \begin{cases} \dfrac{12.5 - p}{12.5} & 0 \leq p \leq 12.5 \\ 0 & p \geq 12.5 \end{cases} \quad [1]$$

$$A_{L}(p) = \begin{cases} \dfrac{p}{12.5} & 0 \leq p \leq 12.5 \\ \dfrac{25 - p}{12.5} & 12.5 \leq p \leq 25 \end{cases} \quad [2]$$

$$A_{M}(p) = \begin{cases} \dfrac{p - 12.5}{25} & 12.5 \leq p \leq 25 \\ \dfrac{37.5 - p}{12.5} & 25 \leq p \leq 37.5 \end{cases} \quad [3]$$

$$A_{H}(p) = \begin{cases} \dfrac{p - 25}{12.5} & 25 \leq p \leq 37.5 \\ \dfrac{50 - p}{12.5} & 37.5 \leq p \leq 50 \end{cases} \quad [4]$$

$$A_{VH}(p) = \begin{cases} \dfrac{p - 37.5}{12.5} & 37.5 \leq p \leq 50 \end{cases} \quad [5]$$

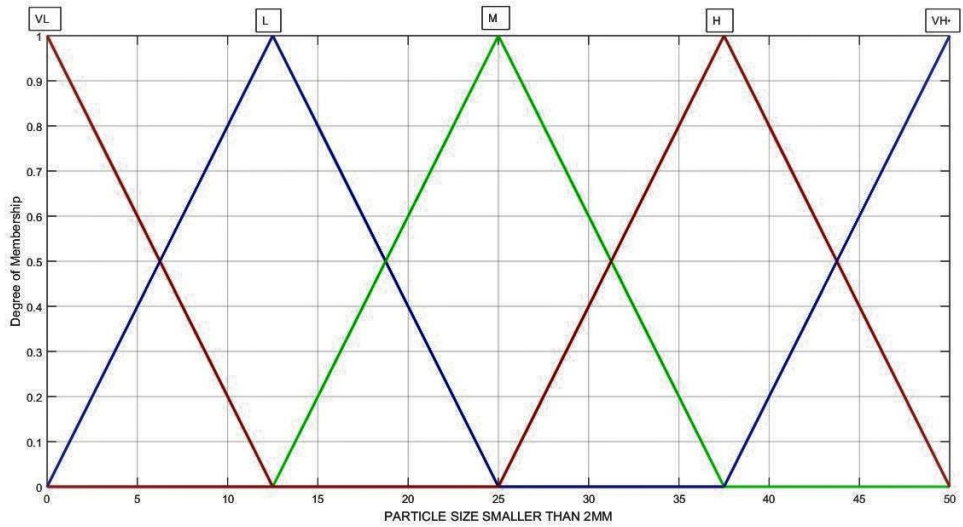

Figure 3. Fuzzy membership function for particle size smaller than 2 mm size

Figure-3 represents the fuzzy membership function developed. The fuzzy descriptor and corresponding membership value for any given particle size smaller than 2mm I S Sieve can be obtained using equation 1 to 5.

### 2.2. Fuzzy Sets and Membership Functions for Coarse Fraction Passing 0.425mm

The membership function for Coarse fraction passing 0.425mm I S Sieve is divided into nine ranges denoted by fuzzy descriptors, namely Very-Very-Low (VVL), Very- Low (VL), Low (L), Low-Medium (LM), Medium (M), Medium-High (MH), High (H) and Very-High (VH).

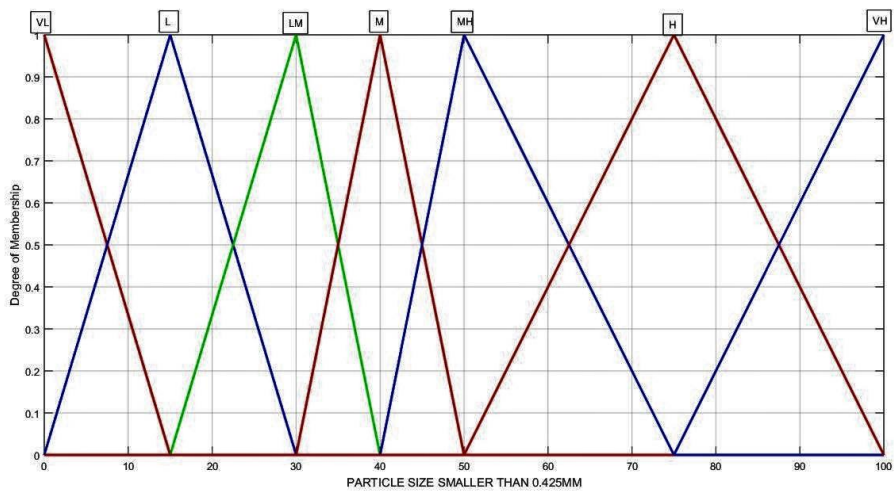

Figure 4. Fuzzy membership function for coarse fraction passing 0.0425 mm

Figure-4 represents the fuzzy membership function developed. The fuzzy descriptor and corresponding membership value for any given coarse fraction smaller than 0.0425mm size can be obtained.

### 2.3. Fuzzy Sets and Membership Functions for Coarse Fraction Passing 0.075mm

The membership function for Coarse fraction passing 0.075mm I S Sieve is divided into eleven ranges denoted by fuzzy descriptors, namely Very-Very-Very Low (VVVL), Very-Very-Low (VVL), Very- Low (VL), Low (L), Low-Medium (LM), Medium (M), Medium-High (MH), High (H) and Very-High (VH), Very- Very- High (VVH), Very-Very- Very- High (VVVH) .

The membership functions are developed for the above fuzzy descriptors.

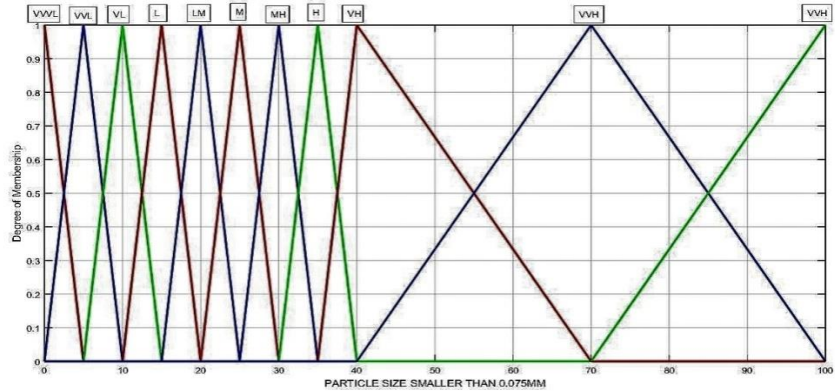

Figure 5. Fuzzy membership function for coarse fraction passing 0.075 mm

Figure-5 represents the fuzzy membership function developed. The fuzzy descriptor and corresponding membership value for any given coarse fraction smaller than 0.075mm size can be obtained.

### 2.4. Fuzzy Sets and Membership Functions for Liquid Limit

The membership function for Liquid Limit is divided into nine ranges denoted by fuzzy descriptors, namely Very-Low (VL), Low (L), Low-Medium (LM), Medium (M), Medium-High (MH), High (H), Very-High (VH) and Very-Very-High (VVH). The membership functions are developed for the above fuzzy descriptors.

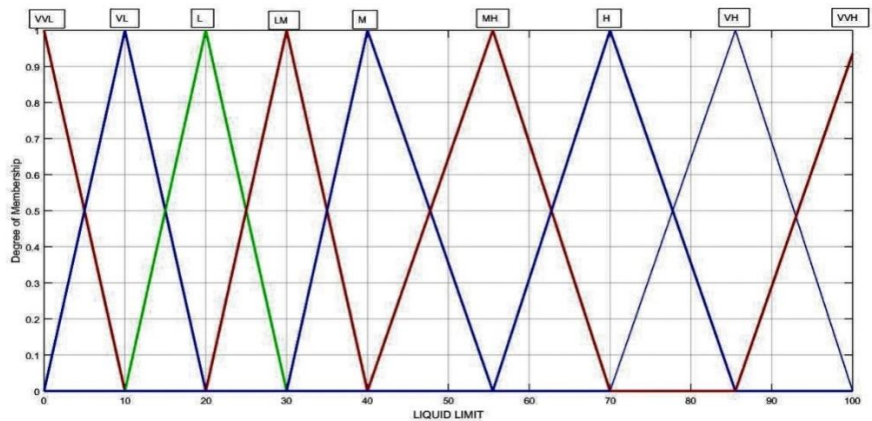

Figure 6.  Fuzzy membership function for Liquid Limit

Figure - 6 represents the fuzzy membership function developed. The fuzzy descriptor and corresponding membership value for any given Liquid limit can be obtained.

## 2.5. Fuzzy Sets and Membership Functions for Plasticity Index

The membership function for Plasticity Index is divided into seven ranges denoted by fuzzy descriptors, namely Very-Low (VL), Low (L), Low-Medium (LM), Medium (M), Medium-High (MH), High (H) and Very -High (VH). The membership functions are developed for the above fuzzy descriptors.

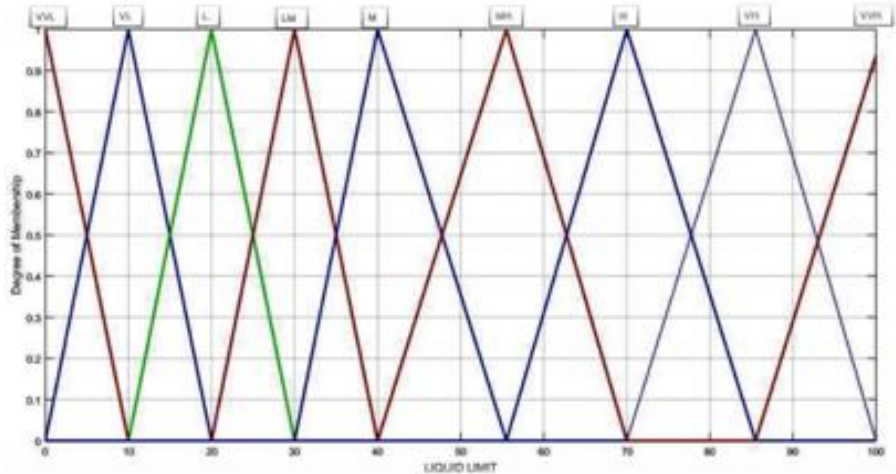

Figure 7. Fuzzy membership function for Plasticity Index

Figure -7 represents the fuzzy membership function developed. The fuzzy descriptor and corresponding membership value for any plasticity index can be obtained.

## 2.6. Fuzzy Rules Generated for Highway Research Board Classification System

The information relating to soil classification is expressed in the form of rules in a Fuzzy rule-based system. Each rule has a set of predecessor propositions consisting of attribute names, namely Particle size smaller than 2mm, Particle size smaller than 0.425mm, Particle size smaller than 0.075mm, liquid limit, plastic limit plasticity Index. The set of rules formed are based on the criteria developed for soil classification as shown in Table 1 [30]. The rules framed for use in this analysis are shown in Table -2. Table 2 displays the rules framed for use in this study. The degree of match for each classification rule shows the classification certainty value. The greater the degree of match the greater the probability of classifying soil into that class.

Table 2: Fuzzy rules constructed for Highway Research Board Classification system

| | |
|---|---|
| $R_1$ | If particle size smaller than 2mm is {VL, L, M, H, VH}, and particle size smaller than 0.425mm is {VL, L}, and Particle size smaller than 0.075mm is {VVVL, VVL, VL}, and Plasticity index is VL, then the soil is classified as A-1-a. |
| $R_2$ | If particle size smaller than 0.425mm is {VL, L, LM, M}, and particle size smaller than 0.075mm is {VVL, VVL, VL, L, M}, and Plasticity index is VL, then the soil is classified as A-1-b. |
| $R_3$ | If particle size smaller than 0.425mm is {H, VH}, and particle size smaller than 0.075mm is {VVL, VVL}, then the soil is classified as A-3. |
| $R_4$ | If particle size smaller than 0.075mm is {VVVL, VVL, VL, L, LM, M, MH}, and liquid limit is {VL, L}, and plasticity index is {VL, L}, then the soil is classified as A-2-4. |
| $R_5$ | If particle size smaller than 0.075mm is {VVVL, VVL, VL, L, LM, M, MH}, and liquid limit is {MH, H, VH, VVH}, and plasticity index is {VL, L}, then the soil is classified as A-2-5. |
| $R_6$ | If particle size smaller than 0.075mm is {VVVL, VVL, VL, L, LM, M, MH}, and liquid limit is {VL, L}, and plasticity index is {M, MH, H, VH}, then the soil is classified as A-2-6. |
| $R_7$ | If particle size smaller than 0.075mm is {VVVL, VVL, VL, L, LM, M, MH}, and liquid limit is {MH, H, VH, VVH} and plasticity index is {M, MH, H, VH}, then the soil is classified as A-2-7. |
| $R_8$ | If particle size smaller than 0.075mm is {VH, VVH, VVVH}, and liquid limit is {VVL, VL, L, LM}, and plasticity index is {VL, L}, then the soil is classified as A-4. |
| $R_9$ | If particle size smaller than 0.075mm is {VH, VVH, VVVH}, and liquid limit is {MH, H, VH, VVH}, and plasticity index is {VL, L}, then the soil is classified as A-5. |
| $R_{10}$ | If particle size smaller than 0.075mm is {VH, VVH, VVVH}, and liquid limit is {VVL, VL, L, LM}, and plasticity index is {M, MH, H, VH}, then the soil is classified as A-6. |
| $R_{11}$ | If particle size smaller than 0.075mm is {VH, VVH, VVVH}, and liquid limit is {MH, H, VH, VVH}, and plasticity index is {M, MH, H, VH}, then the soil is classified as A-7-5/ A-7-6. |

## 3. VALIDATION OF FUZZY RULE BASED MODEL

The proposed Fuzzy rule-based system is considered for soil classification by the Highway Research Board that is implemented in MATLAB. Six soil samples are considered in the proposed work for validation of the Fuzzy rule-based classification developed. The laboratory tests result of soil samples are as shown in Table 3. The soil's index properties are fuzzified and their membership values are obtained from Figure 3-7 for the samples1 to 6 are shown in Table 4-8. We use the definitive fuzzy rules developed to identify the samples, and the degree of possibility for the output attributes are shown in Table 9.

Table 3: Soil Index Testing Summary

| Index properties of Soil | Soil Sample – 1 | Soil Sample - 2 | Soil Sample – 3 | Soil Sample – 4 | Soil Sample - 5 | Soil Sample - 6 |
|---|---|---|---|---|---|---|
| Particle size smaller than 4mm | 100 | 100 | 100 | 100 | 100 | 38 |
| Particle size smaller than 0.425mm | 100 | 80 | 100 | 76 | 100 | 30 |
| Particle size smaller than 0.075mm | 30 | 40 | 92 | 07 | 78 | 11 |
| Liquid Limit | 32 | 25 | 65 | 19 | 34 | 23 |
| Plastic Limit | 21 | 17 | 25 | 16 | 10 | 19 |

Table 4: Membership Values corresponding to Particle size smaller than 2mm size

| Soil Samples | VL | L | M | H | VH |
|---|---|---|---|---|---|
| 1 | 0 | 0 | 0 | 0 | 0 |
| 2 | 0 | 0 | 0 | 0 | 0 |
| 3 | 0 | 0 | 0 | 0 | 0 |
| 4 | 0 | 0 | 0 | 0 | 0 |
| 5 | 0 | 0 | 0 | 0 | 0 |
| 6 | 0 | 0 | 0 | 0.96 | 0.04 |

Table 5: Membership Values corresponding to particle size smaller than 0.0425mm size

| Soil Samples | VL | L | LM | M | MH | H | VH |
|---|---|---|---|---|---|---|---|
| 1 | 0 | 0 | 0 | 0 | 0 | 0 | 1 |
| 2 | 0 | 0 | 0 | 0 | 0 | 0.8 | 0.2 |
| 3 | 0 | 0 | 0 | 0 | 0 | 0 | 1 |
| 4 | 0 | 0 | 0 | 0 | 0 | 0.96 | 0.04 |
| 5 | 0 | 0 | 0 | 0 | 0 | 0 | 1 |
| 6 | 0 | 0 | 1 | 0 | 0 | 0 | 0 |

Table 6: Membership function corresponding to particle size corresponding to 0.075mm size

| Samples | VVVL | VVL | VL | L | LM | M | MH | H | VH | VVH | VVVH |
|---|---|---|---|---|---|---|---|---|---|---|---|
| 1 | 0 | 0 | 0 | 0 | 0 | 0 | 1 | 0 | 0 | 0 | 0 |
| 2 | 0 | 0 | 0 | 0 | 0 | 0 | 0 | 0 | 1 | 0 | 0 |
| 3 | 0 | 0 | 0 | 0 | 0 | 0 | 0 | 0 | 0 | 0.2667 | 0.7333 |
| 4 | 0 | 0.6 | 0.4 | 0 | 0 | 0 | 0 | 0 | 0 | 0 | 0 |
| 5 | 0 | 0 | 0 | 0 | 0 | 0 | 0 | 0 | 0 | 0.7333 | 0.2667 |
| 6 | 0 | 0 | 0.8 | 0.2 | 0 | 0 | 0 | 0 | 0 | 0 | 0 |

Table7: Membership function corresponding to liquid limit

| Samples | VVL | VL | L | LM | M | MH | H | VH | VVH |
|---|---|---|---|---|---|---|---|---|---|
| 1 | 0 | 0 | 0 | 0.8 | 0.2 | 0 | 0 | 0 | 0 |
| 2 | 0 | 0 | 0.5 | 0.5 | 0 | 0 | 0 | 0 | 0 |
| 3 | 0 | 0 | 0 | 0 | 0 | 0.3333 | 0.6667 | 0 | 0 |
| 4 | 0 | 0.1 | 0.9 | 0 | 0 | 0 | 0 | 0 | 0 |
| 5 | 0 | 0 | 0 | 0.6 | 0.4 | 0 | 0 | 0 | 0 |
| 6 | 0 | 0 | 0.7 | 0.3 | 0 | 0 | 0 | 0 | 0 |

Table8: Membership function corresponding to Plasticity Index

| Samples | VL | L | LM | M | MH | H | VH |
|---|---|---|---|---|---|---|---|
| 1 | 0 | 0 | 0.2667 | 0.7333 | 0 | 0 | 0 |
| 2 | 0 | 0 | 0.5333 | 0.4667 | 0 | 0 | 0 |
| 3 | 0 | 0 | 0 | 1 | 0 | 0 | 0 |
| 4 | 0 | 0 | 0.6 | 0.4 | 0 | 0 | 0 |
| 5 | 0 | 0 | 1 | 0 | 0 | 0 | 0 |
| 6 | 0 | 0 | 0.4 | 0.6 | 0 | 0 | 0 |

Table 9: Degree of Possibility of each subgroups of HRB Classification system

| Soil Subgroup | Soil Sample-1 | Soil Sample-2 | Soil Sample-3 | Soil Sample-4 | Soil Sample-5 | Soil Sample-6 |
|---|---|---|---|---|---|---|
| A-1-a | 0 | 0 | 0 | 0.11 | 0 | 0.9187 |
| A-1-b | 0 | 0 | 0 | 0.53 | 0 | 0.8067 |
| A-3 | 0.1 | 0.56 | 0.1 | 0.9240 | 0.1 | 0.0 |
| A-2-4 | 0.88 | 0.1 | 0 | 0.62 | 0.06 | 0.7767 |
| A-2-5 | 0.8 | 0.05 | 0.0667 | 0.53 | 0 | 0.7067 |
| A-2-6 | 0.9733 | 0.05 | 0.1 | 0.57 | 0.1533 | 0.71 |
| A-2-7 | 0.8933 | 0 | 0.1667 | 0.48 | 0.0933 | 0.64 |
| A-4 | 0.08 | 0.9 | 0.5867 | 0.14 | 0.6467 | 0.1367 |
| A-5 | 0 | 0.85 | 0.6533 | 0.05 | 0.5867 | 0.0667 |
| A-6 | 0.8267 | 0.15 | 0.8733 | 0.09 | 0.88 | 0.07 |
| A-7-5 | 0.7467 | 0.1 | 0.94 | 0 | 0.82 | 0 |

**The Illustration of the Proposed fuzzy Expert System is as Shown:**

The input parameters are fuzzified in terms of fuzzy linguistic variables and the corresponding membership values are calculated. The generated definitive fuzzy rules are applied for the classification of above soil samples and the degree of possibility are tabulated. From Table 9, for soil sample1, we notice that fuzzy rule $R_7$ obtains the highest degree of possibility ($R_7$= 0.9733) amidst the fuzzy rules $R_1, R_2, R_3, .................R_{11}$. Therefore, Sample 1 is classified as A-2-6. Usual types of considerable component materials are silty or clayey gravel and sand and the general grading as subgrade is fair to poor. This classification result coincides with the laboratory classification system.

The analysis is repeated for sample 2, 3, 4, 5 and 6. For sample 2, $R_9 = 0.85$ has the highest degree of possibility indicating the soil as A-4 with usual types of considerable component materials are silty soil and the general grading as subgrade is fair to poor . For sample 3, $R_{11} = 0.94$ has the highest degree of possibility indicating the soil as A-7-6 with usual types of considerable component materials are clayey soils and the general grading as subgrade is fair to poor. For sample 4, $R_3 = 0.9240$ has the highest degree of possibility indicating the soil as A-3 with usual types of considerable component materials are fine sand and the general grading as subgrade is excellent to good. For sample 5, $R_{10} = 0.88$ has the highest degree of possibility indicating the soil as A-6 with usual types of considerable component materials are clayey soil and the general grading as subgrade is fair to poor. For sample 6, $R_1 = 0.9187$ has the highest degree of possibility indicating the soil as A-1-a with usual types of considerable component materials are gravel and sand and the general grading as subgrade is excellent to good.

## 4. CONCLUSION

Mathematical models in deterministic form are used to solve the qualitative problems in engineering. But there are uncertainties due to complex nature of problem. In this paper, fuzzy rule-based model is developed in MATLAB by defining fuzzy sets for the index properties of soils, and 11 definitive fuzzy rules are proposed to quantify Highway Research Board classification of soil in qualitative terms. The fuzzy membership functions and fuzzy rules have been defined using statistical data from previous studies that have defined and analysed the different object-oriented metrics. Six soil samples are considered for the validation of the developed model. Soil sample 1 is classified as A-2-6, and its rating as subgrade material is fair to poor. Soil sample 2 is classified as A-4, and its rating as subgrade material is fair to poor. Soil sample 3 is classified as A-7-6 and its rating as subgrade material is fair to poor. Soil sample 4 is classified as A-3, and its rating as subgrade material is excellent to good. Soil sample 5 is classified as A-6, and its rating as subgrade material is fair to poor. Soil sample 6 is classified as A-1-a, and its rating as subgrade material is excellent to poor. The results obtained by using the fuzzy rule-based system coincide with the laboratory test results. This indicates that the developed fuzzy rule-based system can be effectively used for Highway research board classification of soil.